\pdfoutput=1

\documentclass[11pt]{article}

\usepackage[]{acl}

\usepackage{times}
\usepackage{latexsym}
\usepackage{multirow}
\usepackage{amsmath}
\usepackage{xcolor}
\usepackage{amssymb}
\usepackage{arydshln}
\usepackage[T1]{fontenc}

\usepackage[utf8]{inputenc}
\usepackage{comment}
\usepackage{microtype}

\usepackage{inconsolata}
\usepackage{subcaption}
\usepackage{graphicx}
\usepackage{tikz}
\newcommand{\redcross}{\tikz\draw[red, thick] (0,0) -- (0.2,0.2) (0,0.2) -- (0.2,0);}
\newcommand{\greencheck}{\tikz\draw[green, thick] (0,0) -- (0.1,-0.1) -- (0.2,0.125);}
\usepackage{xcolor}
\definecolor{red}{rgb}{0.99, 0.02, 0.02}
\NewDocumentCommand{\heng}
{ mO{} }{\textcolor{red}{\textsuperscript{\textit{Heng}}\textsf{\textbf{\small[#1]}}}}

\title{\raisebox{-0.2em}{\includegraphics[height=1.2em]{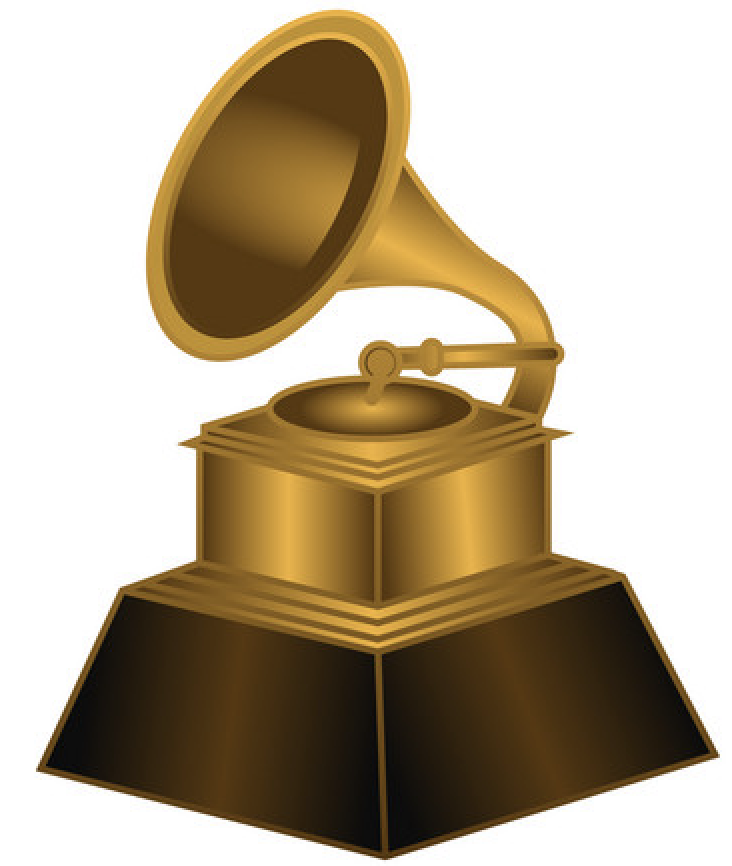}} \textsc{AGRaME}: Any-Granularity Ranking with Multi-Vector Embeddings}

\author{Revanth Gangi Reddy$^1$\thanks{Equal Contribution. Revanth is an external collaborator.}\hspace{1em}Omar Attia$^2$\footnotemark[1]\hspace{1em}Yunyao Li$^3$\thanks{Work done during position at Apple.} \hspace{1em}Heng Ji$^1$\hspace{1em}Saloni Potdar$^2$\\
$^1$University of Illinois at Urbana-Champaign \hspace{1em}  $^2$Apple \hspace{1em} $^3$Adobe\\
  \texttt{\{revanth3,hengji\}@illinois.edu}\\ \texttt{\{oattia,s\_potdar\}@apple.com} \hspace{1em} \texttt{yunyaol@adobe.com}\\
  }

\begin{document}
\maketitle
\begin{abstract}
Ranking is a fundamental and popular problem in search. However, existing ranking algorithms usually restrict the granularity of ranking to full passages or require a specific dense index for each desired level of granularity. Such lack of flexibility in granularity negatively affects many applications that can benefit from more granular ranking, such as sentence-level ranking for open-domain question-answering, or proposition-level ranking for attribution. In this work, we introduce the idea of \textit{any-granularity ranking} which leverages multi-vector approaches to rank at varying levels of granularity while maintaining encoding at a single (coarser) level of granularity. We propose a multi-granular contrastive loss for training multi-vector approaches, and validate its utility with both sentences and propositions as ranking units. Finally, we demonstrate the application of proposition-level ranking to post-hoc citation addition in retrieval-augmented generation, surpassing the performance of prompt-driven citation generation. 
\end{abstract}

\section{Introduction}

\begin{figure}[t]
    \centering
    \includegraphics[width=1.0\linewidth]{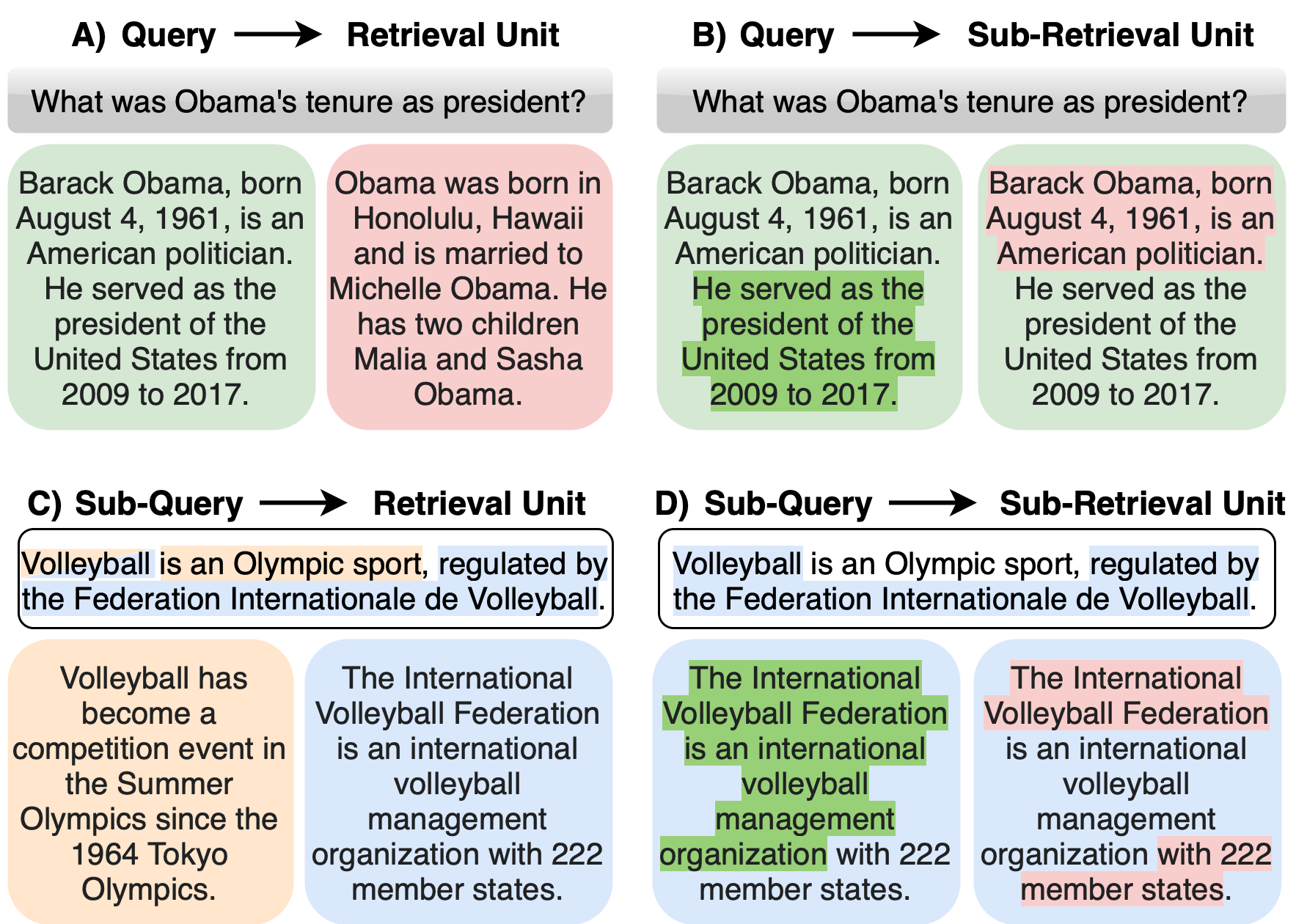}
    \vspace{-1em}
    \caption{Ranking at different levels of granularity. \textit{X}$\rightarrow$\textit{Y} is used to denote that \textit{X} represents the query granularity used for ranking, with entire query encoded, and \textit{Y} indicates the granularity of the retrieval unit being ranked, with entire retrieval unit encoded. In addition to the typical ranking setting (\textit{A}), our proposed approach enables ranking finer retrieval units (\textit{B} and \textit{D}) or using finer query units for ranking (\textit{C} and \textit{D}).}
    
    \label{fig:any_granularity}
\end{figure}

Dense Retrieval approaches leverage dual encoder models to obtain vector representations for both queries and passages. Commonly, single-vector methods~\cite{gautier2022unsupervised, karpukhin2020dense} obtain a single embedding for each query and passage to compute the relevance score using a dot product. In contrast, multi-vector approaches~\cite{khattab2020colbert, santhanam2022colbertv2} capture more fine-grained interactions when computing the query-passage relevance score, resulting in better ranking performance~\cite{thakur2021beir}.
A key advantage of multi-vector approaches is their use of token-level embeddings paired with a MaxSim operation~\cite{khattab2020colbert} for relevance scoring. This enables a more granular scoring mechanism that involes computing dot products between each query token embedding and each passage token embedding, to identify the best matching passage token for each query token. These individual token-level matching scores are subsequently aggregated to obtain the final query-passage relevance score.


We make the important observation that the use of token-level embeddings in multi-vector approaches can facilitate discriminative scoring of different sub-parts within a retrieval unit. For example, even when passages (retrieval units) are input to the encoder, i.e. encoding at passage-level, sub-components such as sentences can be separately scored against the query to identify the most relevant sentence \textit{within} the passage. We argue that such finer-granularity scoring is inherent to multi-vector approaches, but not possible with single-vector approaches, where only one embedding represents the entire passage, thereby not allowing for scoring of its constituent sentences. Being able to rank at varying levels of granularity is beneficial for a variety of applications. For instance, in open-domain question answering~\cite{lee2019latent, karpukhin2020dense}, ranking sentences \textit{within} the retrieved passages can better pinpoint the answer. For attribution~\cite{rashkin2023measuring, chen2023propsegment}, atomic facts (propositions) within sentences need to be used as queries to obtain relevant evidence supporting the facts.

To achieve this, we introduce \textsc{AGRaME} (\textbf{A}ny-\textbf{G}ranularity \textbf{Ra}nking with \textbf{M}ulti-vector \textbf{E}mbeddings), a method that permits ranking at different levels of granularity while maintaining encoding at a single, coarser level. Our approach enables i) ranking at a finer level than the retrieval unit, and ii) ranking using fragments of the query, as demonstrated in Figure \ref{fig:any_granularity}. We hypothesize that encoding at a coarser level--such as the entire retrieval unit or query--can provide additional context for the sub-retrieval units being ranked or sub-parts of the query used for ranking. In contrast, achieving such granularity with single-vector approaches requires the use of specialized encoders, such as a sub-sentence encoder~\cite{chen2023sub}, or necessitates a separate encoding at the desired ranking granularity~\cite{chen2023dense}.

Firstly, how well do multi-vector approaches perform when used for ranking at a finer granularity? To investigate this, we conduct an exploratory experiment (described in \S{\ref{sec:motivating_exp}}) using ColBERTv2~\cite{santhanam2022colbertv2}, a popular multi-vector model, to rank both sentences and passages when encoded at corresponding granularities. From the results summarized in Table \ref{tab:motivating_exp}, we see that performance of sentence ranking is notably inferior when the encoding is at the passage-level, a result that counter-intuitive as passage-level encoding should provide a richer context for scoring sentences.



To improve the model's ability to rank at finer granularity, we propose a multi-granular contrastive loss during training (outlined in \S{\ref{sec:joint_training}). This introduces an additional sentence-level ranking loss that augments a passagr-level loss, to enable the model to not only accurately select the relevant passage for a query but also discriminatively identify the right sentence within that passage. Our experimental results, presented in Section \ref{sec:query_sentence}, confirm significantly boost in sentence-level ranking while maintaining passage-level performance.


While \textsc{AGRaME} is generally applicable to arbitrary granularity, we explore the effectiveness for proposition-level ranking, crucial for applications requiring fine-grained attribution~\cite{rashkin2023measuring}. Our results (in \S{\ref{sec:sub_query_sub_sentence}}) indicate that incorporating a sentence-level contrastive loss further improves ranking at proposition-level. Additionally, we demonstrate (in \S{\ref{sec:citation_addition}}) that proposition-level ranking can effectively integrate citations into generated text post-hoc. Our proposed \textsc{PropCite} method utilizes propositions from generated text as queries to rank input context passages and select relevant citations, showing superior performance over traditional methods that prompt models to include citations in retrieval-augmented generation.

The main contributions are as follows:
\begin{itemize}
    \item We introduce \textsc{AGRaME}, that leverages multi-vector embeddings for ranking at various granularities while using the same encoding-level.
    \item We introduce a multi-granular contrastive loss for training multi-vector approaches, which we show improves sentence-level ranking even when encoding at passage-level.
    \item  We demonstrate superior proposition-level ranking using \textsc{AGRaME}, surpassing existing state-of-the-art methods.
    \item We leverage proposition-level ranking to formulate a post-hoc citation addition approach for retrieval-augmented generation, that outperforms prompt-driven citation generation.
\end{itemize}

\section{Motivating Experiment}
\label{sec:motivating_exp}

Here, we investigate the effectiveness of ColBERTv2~\cite{santhanam2022colbertv2}, a multi-vector approach, in ranking at a finer granularity than the encoding level. Specifically, when encoding is at passage-level, we measure the performance while ranking at sentence level (using the scoring scheme described in \S{\ref{sec:subunit_scoring}}), in addition to ranking at the usual passage-level. A MaxSim operation is applied between query token vectors and token vectors corresponding to the sentence to get a sentence-level score, which is then added with the passage-level score to get the final query-sentence relevance score for ranking. When encoding is at sentence-level, the usual MaxSim score gives query-sentence relevance. On the other hand, the query-passage relevance score for ranking, when encoding is at sentence-level, is obtained as the maximum of the corresponding passage's query-sentence relevance scores. 

We also include Contriever~\cite{gautier2022unsupervised}, a single-vector approach, for comparison. When encoding is at sentence-level, the same strategy as described before is used to obtain query-passage relevance score for Contriever. On the other hand, when encoding is at passage-level, Contriever does not support sentence-level ranking, which as discussed earlier, is an inherent limitation of single-vector approaches. For the evaluation setting, we consider the development set of the Natural Questions~\cite{kwiatkowski2019natural} dataset to get the queries, and Wikipedia\footnote{We use the 22M passage split of the Wikipedia 2018 dump from \citet{gautier2022unsupervised}} as the retrieval corpus. To keep the retrieval index size manageable, Contriever is used to index and retrieve 100 passages, which are then ranked by ColBERTv2. When ranking or encoding at sentence-level, only the sentences in these top 100 passages are considered. Evaluation is done for Precision@1 and Recall@5 based on string exact-match of the answer~\cite{rajpurkar2016squad}.

\begin{table}[t]
    \centering
    \setlength{\tabcolsep}{0.385em}
    \def\arraystretch{1.3}
    \small
    \begin{tabular}{c|c|cc|cc}
    \multirow{3}{*}{\textbf{Model}} & \multirow{3}{*}{\parbox{4em}{\centering \textbf{Encoding Level}}} & \multicolumn{4}{c}{\textbf{Ranking Level}} \\
    & & \multicolumn{2}{c}{\textbf{Sentence}}& \multicolumn{2}{c}{\textbf{Passage}} \\
    & & \textbf{P@1} & \textbf{R@5} & \textbf{P@1} & \textbf{R@5} \\
    \hline
    \multirow{2}{*}{\parbox{6em}{\centering Contriever (Single Vec.)}} & Sentence & 19.3 & 45.6 & 32.4 & 62.8\\
    & Passage & - & - & 37.8 & 65.1 \\
    \hline
    \multirow{2}{*}{\parbox{6em}{\centering ColBERTv2 (Multi Vec.)}} & Sentence & \textbf{31.6} & \textbf{56.3} & 40.2 & 66.8\\
    & Passage & 27.4 & 48.8 & \textbf{43.4} & \textbf{69.1}\\
    \hline
    \end{tabular}
    \caption{Precision@1 (P@1) and Recall@5 (R@5) results on the Natural Questions~\cite{kwiatkowski2019natural} dev set. We show numbers both at sentence-level and passage-level ranking granularities for when sentences and passages are encoded individually.}
    \label{tab:motivating_exp}
\end{table}

Table \ref{tab:motivating_exp} shows the results, wherein we see a substantial drop in sentence-level ranking when encoding is at passage-level, and vice versa. As expected, passage-level ranking is better when encoding is at passage-level. However, it is surprising to see sentence-level ranking performance to be lower since passage-level encoding can provide more context when encoding the individual tokens in the sentences. This is more so the case when sentences in the passage that actually `answer' the query do not have any overlapping terms (semantic or lexical) with the query. Table \ref{tab:motivating_example} shows an example for this. At both sentence and passage-level encoding, sentences S1 and S2, on account of strong lexical overlap with the query, are scored considerably higher than sentence S3, which actually describes the effects of climate change but has weak semantic overlap with the query. From the token-wise MaxSim score heatmap, we can see that tokens in S3, the most relevant sentence, get lower scores. 

We can expect scoring S1 and S2 highly, on account of overlap, to be particularly useful when identifying the passage as relevant amongst a corpus of millions of passages. On the other hand, it can be counter-productive when discriminatively selecting the most relevant sentence \textit{within} the passage.
 This suggests that ranking at different granularities requires the model to have the ability to dynamically switch the notion of relevance when scoring. As we describe later in \S{\ref{sec:subunit_scoring}}, we introduce a new query marker during encoding 
 to signal the level of granularity, which helps the model distinguish better when scoring at different granularities  
 In \S{\ref{sec:query_sentence}}, we demonstrate that this helps to score appropriately for ranking at sentence-level (a finer granularity), when encoding at passage-level.

\begin{table}[t]
    \centering
    \setlength{\tabcolsep}{0.4em}
\def\arraystretch{1.3}
    \small
    \begin{minipage}{0.95\linewidth}
     \includegraphics[width=1.0\linewidth]{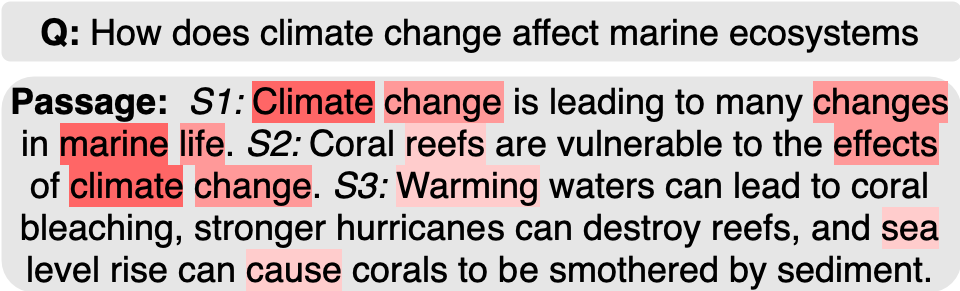}
    \end{minipage}
    \begin{tabular}{c|c|c}    
    \hline
    \hline
    \textbf{Sent. ID} & \textbf{Sentence-level Enc.} & \textbf{Passage-level Enc.} \\
    \hline
    S1 \redcross  & Rank:\textbf{1}, Score:\textbf{23.31} & Rank:\textbf{1}, Score:\textbf{23.92}\\
    S2 \redcross & Rank:2, Score:17.47 & Rank:2, Score:20.12\\
    S3 \greencheck & Rank:3, Score:16.63 & Rank:3, Score:16.96\\
    \hline
    \end{tabular}
    \caption{Sentence-level ColBERTv2 scores for different sentences in the same passage, when encoding is at sentence-level and passage-level. We see that the most relevant sentence (S3) is ranked worst (i.e lowest score). Token-wise MaxSim score heatmap is also shown, with tokens in S1 and S2 having higher scores than in S3.} 
    \label{tab:motivating_example}
\end{table}

\section{Method}

Here, we first provide background (in \S{\ref{sec:colbertv2}}) on the modeling and training process for ColBERTv2~\cite{santhanam2022colbertv2}. Then, we describe our proposed multi-granular contrastive training process (in \S{\ref{sec:joint_training}}), which provides an additional sentence-level relevance supervision during distillation.

\subsection{ColBERTv2 Preliminaries}
\label{sec:colbertv2}
Single-vector retrievers~\cite{karpukhin2020dense, gautier2022unsupervised} typically use a BERT-based~\cite{devlin2019bert} dual-encoder architecture to obtain a single embedding for a query $q$ and passage $p$ separately. This is usually the \textit{CLS} output or a pooled representation of the individual token outputs from the encoder $E(.)$. The query-passage relevance score is computed as the dot product of their corresponding representation:
\begin{equation*}
    \vec{Q_q} = Pool(E(q)); \vec{P_p} = Pool(E(p))
\end{equation*}  
 \begin{equation*}
    Score(q,p) = \vec{Q_q}^T\vec{P_p}
 \end{equation*}
 In constrast, ColBERTv2~\cite{santhanam2022colbertv2} is a multi-vector retrieval model, that uses token-level dense embeddings of the query and passage. Given a query $q$ containing $n$ tokens $t^q_i$ and passage $p$ containing $m$ tokens $t^p_i$, additional query and passage marker tokens $m_q$ and $m_p$ are prepended to the query and passage respectively before encoding, to provide additional signal to the encoder. The query-passage relevance score $S_{CB}(q,p)$ is obtained as below using the \textit{MaxSim} operator introduced in~\citet{khattab2020colbert}:
 \begin{equation*}
     [\vec{Q_{t^q_1}}, \vec{Q_{t^q_1}}, ..., \vec{Q_{t^q_n}}] = E(q) = E(cat(m_q, t^q_1, ..., t^q_n))
 \end{equation*}
  \begin{equation*}
    [\vec{P_{t^p_1}}, \vec{P_{t^p_2}}, ..., \vec{P_{t^p_m}}] = E(p) = E(cat(m_p, t^p_1, ..., t^p_m))
 \end{equation*}
 \begin{equation*}
     S_{CB}(q,p) = MaxSim(q,p) = \sum_{i=1}^{n} \max_{1 \leq j \leq m} \vec{Q_{t^q_i}}^T \vec{P_{t^p_j}}
 \end{equation*}
 The training process for neural retrievers typically involves a contrastive loss over the <query $q$, postitive $p^+$, negative $p^-$> triples. ColBERTv2 instead incorporates a distillation-based training strategy wherein $k$ negative passages are sampled from the retrieval corpus, to form a $(k+1)$-way passage set $[p]=\{p^+,p^-_1, ..., p^-_k\}$ for each query. The relevance supervision is in the form of soft scores $S_{CE}(.)$ from a cross-encoder reranker. A KL-Divergence loss $\mathcal{L}_{psg}$ between the cross-encoder and ColBERT passage scoring distributions, $D_{CE}(q,[p])$ and $D_{CB}(q,[p])$ respectively, is used for training:
 \begin{equation*}
      D_{CB}(q,[p]) = \left[ Softmax(S_{CB}(q,p_i)) \right]_{i=1}^{k+1}
 \end{equation*}
 \begin{equation*}
     D_{CE}(q,[p]) = \left[ Softmax(S_{CE}(q,p_i)) \right]_{i=1}^{k+1}
 \end{equation*}
 \begin{equation*}
 \mathcal{L}_{psg}(q,[p]) = KL(D_{CE}(q,[p]) || D_{CB}(q,[p]))
 \end{equation*}

\subsection{\textsc{AGRaME}: Any-Granularity Ranking with Multi-Vector Embeddings}
\label{sec:subunit_scoring}

\begin{figure}[t]
    \centering
    \includegraphics[width=1.0\linewidth]{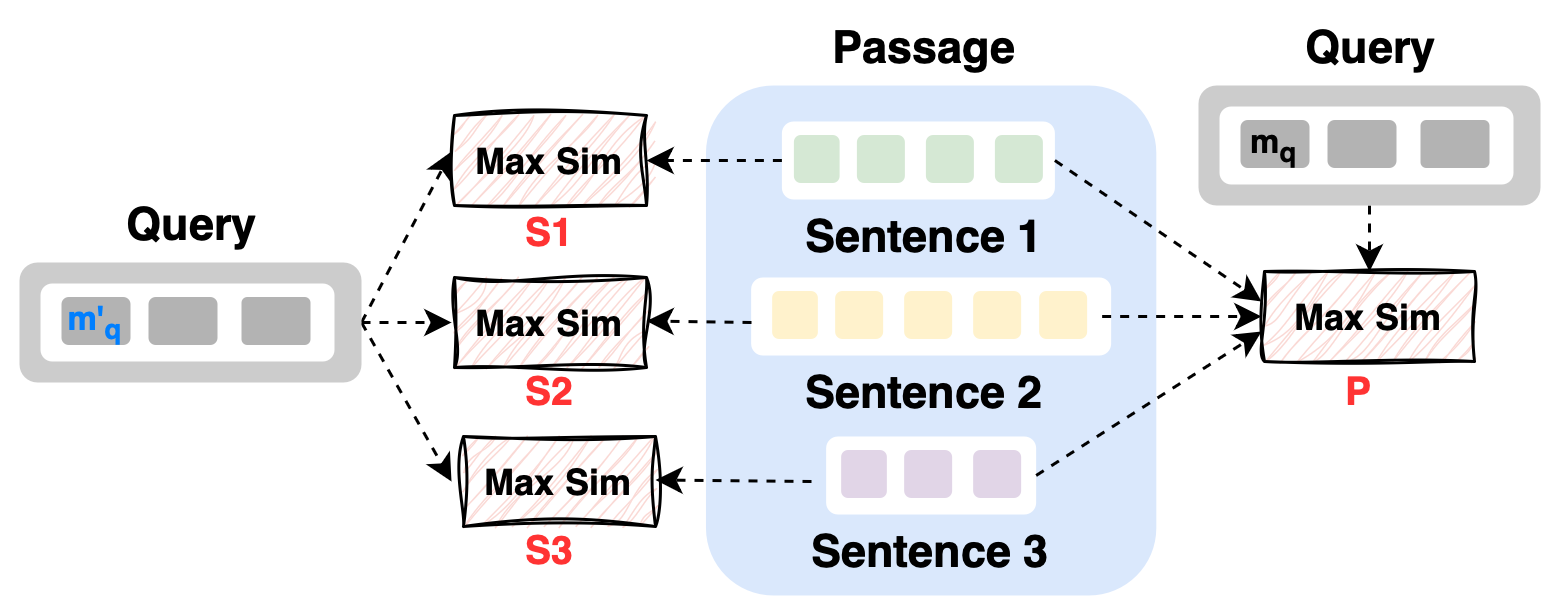}
    \vspace{-2em}
    \caption{Figure demonstrating our sentence-level scoring methodology using multi-vector representations with encoding at passage-level. Query marker $m_q$ is used while getting  passage-level score $P$, while marker $m'_q$ is used for getting sentence-level scores $S1$, $S2$, $S3$.}
    \label{fig:sentence_scoring}
\end{figure}

Here, we introduce our approach for scoring sub-units within the retrieval unit. This is made possible by the access to token-level embeddings in multi-vector approaches. While \textsc{AGRaME} can rank at any granularity, in this section, we will consider sentences as the sub-units for simplicity. With the entire passage input to the encoder, only the output embeddings corresponding to tokens within a given sentence are used during the \textit{MaxSim} operation for scoring that sentence. 


Let $t^{p_i}_{jr}$ correspond to the $j^{th}$ token of sentence $s_j^{p_i}$ from passage $p_i$ that is passed as input to encoder $E$. To signal the model to score discriminatively \textit{within the passage} for sentence-level ranking, we prepend a new query marker token $m'_q$, different from $m_q$ used when ranking at passage-level. The \textit{in-passage} query-sentence relevance score $S_{CB}(q,s_j^{p_i})$ is computed as follows:
\begin{equation*}
     [\vec{Q'_{t^q_1}}, \vec{Q'_{t^q_1}}, ..., \vec{Q'_{t^q_n}}] = E(cat(m'_q, t^q_1, ..., t^q_n))
 \end{equation*}
\begin{equation*}
    S_{CB}(q,s_j^{p_i}) = \sum_{i=1}^{n} \max_{1 \leq r \leq |s_j^{p_i}|} \vec{Q'_{t^q_i}}^T \vec{P_{t^{p_i}_{jr}}}
\end{equation*}
Note that the passage encoding is the same as before, meaning the same multi-vector index can be used for both passage-level and sentence-level ranking. As we demonstrate in \S{\ref{sec:query_sentence}}, encoding at passage-level provides more context to the token embeddings to benefit sentence-level ranking.

 We note that our proposed sentence-level loss (described in \S{\ref{sec:joint_training}}) teaches the model to rank sentences discriminatively \textit{within a passage}, and not \textit{across passages}. Hence, at inference to get a final sentence-level relevance score $Score(q,s_j^{p_i})$ to rank sentences across passages, we combine the in-passage sentence relevance score $S_{CB}(q,s_j^{p_i})$ with the usual passage-level relevance score $S_{CB}(q,p_i)$:
 \begin{equation*}
     Score(q,s_j^{p_i}) = S_{CB}(q,s_j^{p_i}) + \alpha S_{CB}(q,p_i)
 \end{equation*}

\subsection{Multi-Granular Contrastive Training}
\label{sec:joint_training}
As discussed in \S{\ref{sec:colbertv2}}, given a query $q$ and a passage set $[p]$, the ColBERTv2 training process aims to teach the model to identify the most relevant passage within $[p]$. To enable the model to discriminatively select sub-units within the passage, we propose to incorporate a more finer-level of training supervision, by teaching to further identify the most relevant sentence within each passage. 

Since ColBERTv2 uses passage-level cross-encoder scores as teacher supervision, we train a different cross-encoder model $CE'$ to provide in-passage sentence-level relevance supervision. Specifically, $CE'$ takes a passage $p_i$ as input, with a given sentence $s_j^{p_i}$ marked with delimiters \$, to give a relevance score $S_{CE'}(q,s_j^{p_i})$ for the sentence. 

\begin{equation*}
    S_{CE'}(q,s_j^{p_i}) = CE'(q,cat(s_1^{p_i},...,\$s_j^{p_i}\$,...,s_l^{p_i}))
\end{equation*}
$CE'$ is trained using question answering data in the form <query, passage, answer> triples. A binary cross-entropy loss is used while training $CE'$, wherein any sentence within the passage that contains the answer is marked as a positive, with the other sentences marked as negatives. 

The cross encoder $CE'$ provides soft scores for sentence-level relevance superivision when training our model. For each passage $p_i$, we compute a KL-divergence loss $\mathcal{L}_{s}(q,p_i)$ between the $CE'$ and ColBERTv2 sentence-level scoring distributions, $D_{CE'}(q,[s^{p_i}])$ and $D_{CB}(q,[s^{p_i}])$ respectively.

\begin{equation*}
    D_{CB}(q,[s^{p_i}]) = \left[ Softmax(S_{CB}(q,s_j^{p_i})) \right]_{j=1}^{l}
\end{equation*}
\begin{equation*}
    D_{CE'}(q,[s^{p_i}]) = \left[ Softmax(S_{CE'}(q,s_j^{p_i})) \right]_{j=1}^{l}
\end{equation*}
\begin{equation*}
 \mathcal{L}_{s}(q,p_i) = KL(D_{CE'}(q,[s^{p_i}]) || D_{CB}(q,[s^{p_i}])))
 \end{equation*}

We then aggregate each passage's sentence-level scoring loss $\mathcal{L}_{s}(q,p_i)$, by weighting with the corresponding passage's relevance supervision score $S_{CE}(q,p_i)$, to get a single loss ${L}_{sent.}(q,[p])$. The passage score weight ensures that the model is penalized higher on sentence-level losses for passages that are more relevant. The sentence-level loss ${L}_{sent.}(q,[p])$ is finally added to original passage-level loss ${L}_{psg}(q,[p])$ to get the training loss $\mathcal{L}$. 
\begin{equation*}
 \mathcal{L}_{sent.}(q,[p]) = \sum_{i=1}^{k+1} Softmax(S_{CE}(q,p_i)) \mathcal{L}_{s}(q,p_i)
 \end{equation*}
 \begin{equation*}
 \mathcal{L}(q,[p]) = \mathcal{L}_{psg}(q,[p]) + \mathcal{L}_{sent.}(q,[p])
 \end{equation*}

\section{Experiments}
\label{sec:experiments}


\textsc{AGRaME} can rank at different granularities, as shown in Figure \ref{fig:any_granularity}, which involves ranking sub-parts of the retrieval unit or ranking using sub-parts of the query. In our experiments, we aim to investigate two research questions: \textbf{RQ1:} Can the training approach proposed in \S{\ref{sec:joint_training}} improve ranking at a finer granularity than the level of encoding, i.e. \textit{Query$\rightarrow$Sub-Retrieval Unit}? In \S{\ref{sec:query_sentence}}, we show the improvements at sentence-level ranking from our proposed multi-granular contrastive loss, while maintaining performance at passage-level, i.e.\textit{Query$\rightarrow$Retrieval Unit}; \textbf{RQ2:} Can multi-vector embeddings be used to rank with sub-parts of the query? In \S{\ref{sec:sub_query_sub_sentence}}, we demonstrate the application of multi-vector approaches in \textit{Sub-Query$\rightarrow$Sub-Retrieval Unit} ranking for proposition-level attribution. Here, a given proposition within a sentence is used as the query to rank and identify relevant propositions in a corpus of sentences. Further, in \S{\ref{sec:citation_addition}}, we introduce \textsc{PropCite}, a post-hoc citation addition approach based on \textit{Sub-Query$\rightarrow$Retrieval Unit} ranking. \textsc{PropCite} scores input context passages based on propositions in the generated text to add citations in retrieval-augmented generation.


\subsection{Query$\rightarrow$Sub-Retrieval Unit Ranking for Open-Domain QA}
\label{sec:query_sentence}
In \S{\ref{sec:motivating_exp}}, we saw that with a multi-vector approach (ColBERTv2), sentence ranking performance drops when changing the encoding from sentence-level to passage-level.
We addressed this in two ways: (a) our proposed multi-granular contrastive loss (in \S{\ref{sec:joint_training}}) provides sentence-level relevance supervision at training; b) \textsc{AGRaME} introduces a new query marker (in \S{\ref{sec:subunit_scoring}}) to signal scoring at sentence-level.
In this section, we empirically demonstrate the benefits of our proposed approach by evaluating sentence-level (sub-retrieval unit) ranking performance when encoding is at passage-level.

\begin{table*}[!htb]
    \centering
    \setlength{\tabcolsep}{0.385em}
    \def\arraystretch{1.3}
    \scriptsize
    \begin{tabular}{c|c|cc|cc|cc|cc|cc|cc|cc|cc}
    \multirow{3}{*}{\textbf{Model}} & \multirow{3}{*}{\parbox{4em}{\centering \textbf{Encoding Level}}} & \multicolumn{4}{c|}{\textbf{Natural Questions}} & \multicolumn{4}{c|}{\textbf{TriviaQA}} & \multicolumn{4}{c|}{\textbf{Web Questions}} & \multicolumn{4}{c}{\textbf{Entity Questions}}\\
    & & \multicolumn{2}{c}{\textbf{Sentence}}& \multicolumn{2}{c|}{\textbf{Passage}} & \multicolumn{2}{c}{\textbf{Sentence}}& \multicolumn{2}{c|}{\textbf{Passage}} & \multicolumn{2}{c}{\textbf{Sentence}}& \multicolumn{2}{c|}{\textbf{Passage}} & \multicolumn{2}{c}{\textbf{Sentence}}& \multicolumn{2}{c}{\textbf{Passage}} \\
    & & \textbf{P@1} & \textbf{R@5} & \textbf{P@1} & \textbf{R@5} & \textbf{P@1} & \textbf{R@5} & \textbf{P@1} & \textbf{R@5} & \textbf{P@1} & \textbf{R@5} & \textbf{P@1} & \textbf{R@5} & \textbf{P@1} & \textbf{R@5} & \textbf{P@1} & \textbf{R@5}\\
    \hline
    \multirow{2}{*}{Contriever} & Sentence & 20.6 & 48.9 & 35.0 & 65.4 & 31.0 & 58.8 & 48.5 & 72.1 & 14.5 & 39.1 & 28.8 & 57.9 & 14.7 & 42.7 & 39.8 & 64.9 \\
    & Passage & - & - & 40.3 & 66.0 & - & - & 50.1 & 71.5 & - & - & 36.9 & 63.6 & - & - & 36.9 & 63.6\\
    \hline
    \multirow{2}{*}{ColBERTv2} & Sentence & 32.7 & 58.8 & 42.0 & 68.8 & 43.2 & 66.1 & 55.6 & 74.7 & 29.0 & 51.9 & 38.8 & 63.7 & 38.1 & 59.4 & 50.9 & 68.1\\
    & Passage & 27.9 & 51.1 & 43.2 & \textbf{70.0} & 43.5 & 65.6 & 57.5 & \textbf{75.6} & 27.6 & 50.7 & 41.0 & 65.1 & 39.2 & 55.3 & 53.9 & 69.2 \\
    \hline
    Ours & Passage & \textbf{36.8} & \textbf{60.5} & \textbf{44.0} & 69.9 & \textbf{48.9} & \textbf{68.1} & \textbf{57.9} & \textbf{75.6} & \textbf{33.2} & \textbf{55.6} & \textbf{41.2} & \textbf{65.4} & \textbf{43.8} & \textbf{61.5} & \textbf{54.2} & \textbf{69.5} \\
    \hline
    \end{tabular}
    \caption{Precision@1 (P@1) and Recall@5 (R@5) results on various open-domain QA datasets. We show numbers both at sentence-level and passage-level ranking for when sentences and passages are encoded individually.}
    \label{tab:retrieval_openqa}
\end{table*}

\begin{table*}[!htb]
    \centering
    \setlength{\tabcolsep}{0.43em}
    \def\arraystretch{1.3}
    \scriptsize
    \begin{tabular}{c|c|cc|cc|cc|cc|cc|cc|cc||cc}
    \multirow{2}{*}{\textbf{Model}} & \multirow{2}{*}{\parbox{3.9em}{\centering \textbf{Encoding Level}}} & \multicolumn{2}{c|}{\textbf{Finance}} & \multicolumn{2}{c|}{\textbf{Recreation}} & \multicolumn{2}{c|}{\textbf{Lifestyle}} & \multicolumn{2}{c|}{\textbf{Science}} & \multicolumn{2}{c|}{\textbf{Technology}} & \multicolumn{2}{c|}{\textbf{Writing}} & \multicolumn{2}{c||}{\textbf{Biomedical}} & \multicolumn{2}{c}{\textbf{Average}}\\
    & & \textbf{Sent.} & \textbf{Psg.} & \textbf{Sent.} & \textbf{Psg.} & \textbf{Sent.} & \textbf{Psg.} & \textbf{Sent.} & \textbf{Psg.} & \textbf{Sent.} & \textbf{Psg.} & \textbf{Sent.} & \textbf{Psg.} & \textbf{Sent.} & \textbf{Psg.} & \textbf{Sent.} & \textbf{Psg.}\\
    \hline
    \multirow{2}{*}{Contriever} & Sentence & 13.8 & 22.2 & 17.9 & 29.4 & 19.7 & 32.7 & 10.9& 18.8& 11.3& 18.3 & 23.0 &36.1 &10.7 & 16.6 & 15.3 & 24.9\\
    & Passage & - & 27.2& - & 34.7 & - & 40.4 & - & 17.5 & - & 21.4 & - & 39.6 & - & 4.6 & - & 26.5\\
    \hline
    \multirow{2}{*}{ColBERTv2} & Sentence & 15.8 & 23.7 & 24.0 & 33.6 & 22.6 & 34.2 & 17.6 & 25.0 & 15.5 & 23.4 & 33.4 & 46.6 & 12.8 & 17.3 & 20.2 & 29.1\\
    & Passage & 17.1 & \textbf{29.8} & 25.5 & \textbf{40.7} & 23.9 & 41.9 & 18.4 & \textbf{28.7} & 16.7 & \textbf{27.1} & 34.7 & \textbf{51.3} & 13.1 & 16.9 & 21.4 & \textbf{33.8}\\
    \hline
    Ours & Passage & \textbf{19.5} & \textbf{29.8} & \textbf{29.2} & 40.4 & \textbf{30.0} & \textbf{42.6} & \textbf{20.5} & 28.1 & \textbf{18.4} & 26.4 & \textbf{36.7} & 50.2 & \textbf{15.4} & \textbf{17.5} & \textbf{24.2} & 33.6\\
    \hline
    \end{tabular}
    \caption{Precision@1 results on various domains from the RobustQA dataset~\cite{han2023robustqa}. We show numbers at sentence-level and passage-level ranking for when sentences and passages are encoded individually.}
    \label{tab:retrieval_robustqa}
\end{table*}

\subsubsection{Setup}

\paragraph{Datasets}

We first evaluate on different popular open-domain QA datasets: Natural Questions (NQ)~\cite{kwiatkowski2019natural}, TriviaQA~\cite{joshi2017triviaqa}, Web Questions~\cite{berant2013semantic} and Entity Questions~\cite{sciavolino2021simple}. For the retrieval corpus, we use the 2018 Wikipedia dump released by~\citet{lee2019latent}.

For cross-domain evaluation, we consider the RobustQA~\cite{han2023robustqa} dataset, a large-scale OpenQA benchmark specifically designed for evaluating cross-domain generalization capabilities. The QA pairs and documents correspond to various domains like finance (adopted from FiQA~\cite{maia201818}), biomedical (adopted from BioASQ~\cite{tsatsaronis2015overview}), along with recreation, lifestyle, science, technology and writing, which are all adopted from LOTTE~\cite{santhanam2022colbertv2}. 

\paragraph{Baselines}

We use Contriever~\cite{gautier2022unsupervised} as the single-vector baseline, and ColBERTv2~\cite{santhanam2022colbertv2} as the multi-vector baseline. All models use MS MARCO~\cite{nguyen2016ms} as the training dataset. Due to storage constraints, we create a single-vector index with Contriever and rank the top-100 retrieval results from Contriever using the multi-vector approaches to report numbers. 

\subsubsection{Results}

Table \ref{tab:retrieval_openqa} shows ranking results on various open-domain QA datasets. Firstly, as expected, for both Contriever and ColBERTv2, passage-level ranking performance is best when encoding is at passage-level. We observe that our proposed approach significantly improves sentence-level ranking performance with passage-level encoding, even outperforming sentence-level ranking at sentence-level encoding. This result confirms our intuition that passage-level encoding benefits sentence-level ranking, since it can provide more context to the individual sentences during encoding. Moreover, we ensure that passage-level ranking performance is not compromised, with our approach matching that of ColBERTv2 at passage-level encoding. 

Table \ref{tab:retrieval_robustqa} shows sentence-level and passage-level ranking results for cross-domain evaluation on the RobustQA benchmark. 
We observe that our approach is robust and extends to cross-domain settings, with consistent improvements in sentence-level ranking across the board, while passage-level ranking performance almost the same.

\subsubsection{Analysis}

We do an ablation study to analyze the effect of using the new query marker $m'_q$, instead of the default query marker $m_q$, when scoring at sentence-level. Note that the markers $m_q$ and $m'_q$ at inference only affect the query token embeddings. We consider three different settings: A1) Using $m'_q$ for sentence-level ranking at training and inference, which corresponds to our proposed approach, A2) Using $m'_q$ for sentence-level ranking at training but using $m_q$ at inference, and finally A3) Using the default $m_q$ for sentence-level ranking at training and inference. We also include the baseline ColBERTv2 for comparison, which does not have sentence-level supervision at training and uses $m_q$ at inference. 
From the results in Table \ref{tab:query_marker_ablation}, we can see the benefit of using a different query marker, with A1 outperforming A3 in the majority of the cases. Moreover, using $m_q$ at inference even while being trained with $m'_q$ (A2) shows some gains over baseline ColBERTv2, implying that the model also learns to encode passage tokens to be better at discriminatively scoring sentences in the passage. 

In addition, we show the training loss curves in Figure \ref{fig:loss_comparison} when the same query marker ($m_q$) vs different query markers ($m'_q$ and $m_q$) are used for sentence-level and passage-level loss respectively. We can see that the model converges faster at sentence-level loss when new marker $m'_q$ is used. Further, Table \ref{tab:ablation_example} shows the sentence-wise scores for the example in Table \ref{tab:motivating_example} from using $m'_q$ vs $m_q$ for sentence-level scoring. We observe that sentence-level ranking changes when $m'_q$ is used, with the most relevant sentence (S3) ranked best.

\begin{table}[t]
    \centering
    \setlength{\tabcolsep}{0.36em}
    \def\arraystretch{1.3}
    \small
    \begin{tabular}{c|cccc}
    \textbf{Setting}  & \textbf{NQ} & \textbf{TQA} & \textbf{WebQ} & \textbf{EntQ}\\
    \hline
    ColBERTv2 & 27.9 & 43.5 & 27.6 & 39.2\\
    \hline
    A1) Train$\rightarrow$$m'_q$, Rank$\rightarrow$$m'_q$& \textbf{36.8} & \textbf{48.9} & \textbf{33.2} & 43.8\\
    A2) Train$\rightarrow$$m'_q$, Rank$\rightarrow$$m_q$& 29.1 & 44.8 & 29.4 & 40.8 \\
    A3) Train$\rightarrow$$m_q$, Rank$\rightarrow$$m_q$& 35.9 & 47.6 & 32.9 & \textbf{44.1}\\
    
    \hline
    \end{tabular}
    \caption{Precision@1 of sentence-level ranking performance for various variants of using a different query marker. ColBERTv2 is trained only with a passage-level loss and uses the $m_q$ query marker. The latter three variants are represented with the query marker used while training with sentence-level contrastive loss and that used for sentence-level ranking at inference.}
    \label{tab:query_marker_ablation}
\end{table}

\begin{figure}[t]
    \centering
    \begin{subfigure}{0.49\linewidth}       
     \includegraphics[scale=0.1]{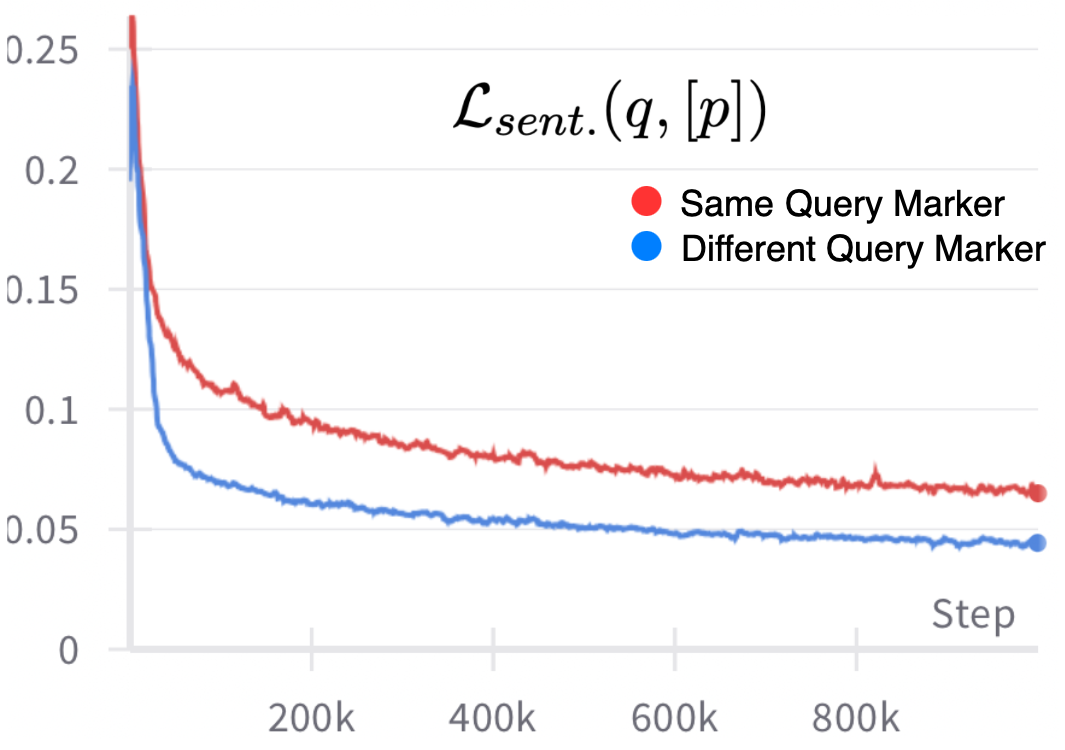}
     \end{subfigure}%
     \hfill
    \begin{subfigure}{0.49\linewidth}       
     \includegraphics[scale=0.1]{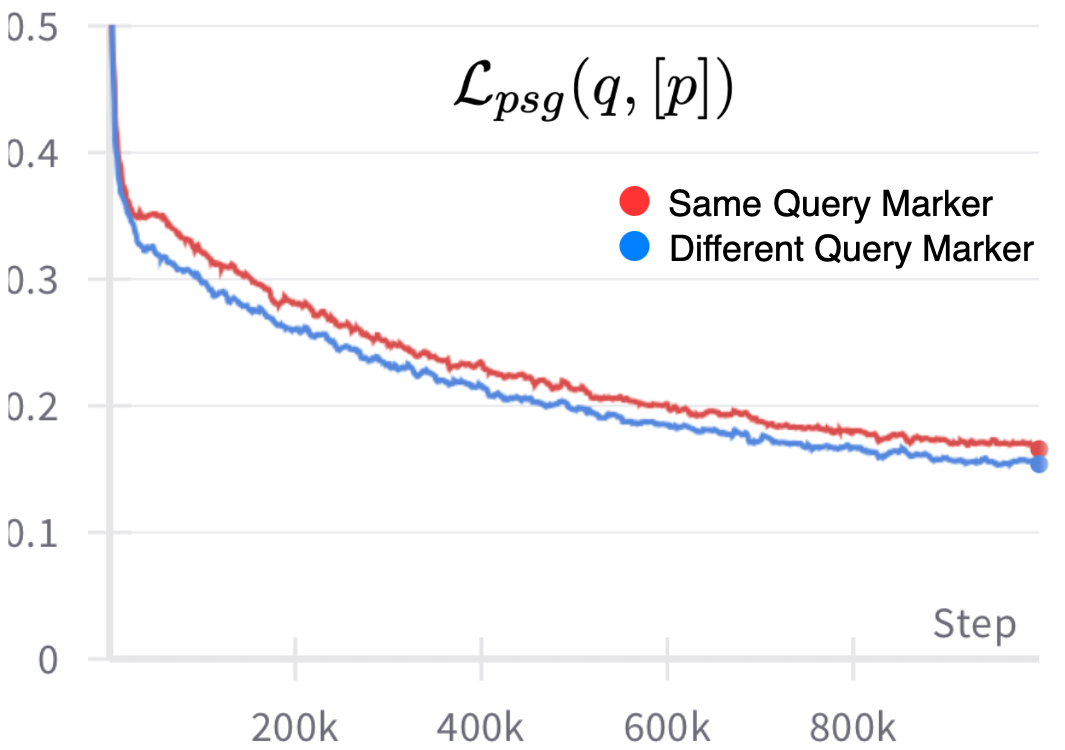}
     \end{subfigure}%
    \caption{Comparison of training curves for sentence-level and passage-level loss, when a different query marker is used. The model converges faster at sentence-level with a different query marker, while passage-level loss is mostly similar for the two.}
    \label{fig:loss_comparison}
\end{figure}

\begin{table}[t]
    \centering
    \setlength{\tabcolsep}{0.4em}
\def\arraystretch{1.3}
    \small
    \begin{tabular}{c|c|c}
    \hline
    \textbf{Sent. ID} & \textbf{Query Marker $m'_q$} & \textbf{Query Marker $m_q$} \\
    \hline
    S1 \redcross & Rank:2, Score:14.32 & Rank:\textbf{1}, Score:\textbf{24.04}\\
    S2 \redcross & Rank:3, Score:14.16 & Rank:2, Score:21.07\\
    S3 \greencheck & Rank:\textbf{1}, Score:\textbf{15.92} & Rank:3, Score:16.81\\
    \hline
    \end{tabular}
    \caption{Sentence-level scores from our model at passage-level encoding for the example in Table \ref{tab:motivating_example}, when different query markers are used. The most relevant sentence (S3) is ranked best when new marker $m'_q$ is used.}
    \label{tab:ablation_example}
\end{table}

\subsection{Sub-Query$\rightarrow$Sub-Retrieval Unit Ranking for Fine-Grained Attribution}
\label{sec:sub_query_sub_sentence}

Attributing model-generated text with supporting information from known sources is an emerging research topic~\cite{gao-etal-2023-rarr, liu2023evaluating}. Each sentence in the generation can have multiple atomic facts or propositions~\cite{min2023factscore} for which evidence needs to be obtained. In this context, we explore ranking at the sub-sentence granularity, wherein given a sentence as a query, fine-grained attributions~\cite{rashkin2023measuring} need to be obtained for a sub-part of the sentence. Specifically, we consider the \textit{Atomic Fact Retrieval} task, wherein given an atomic proposition (sub-query) in the sentence, the system is expected to identify and retrieve evidence from atomic propositions as facts, each of which can be a sub-part of sentences within a corpus.

We consider this task to demonstrate that multi-vector embeddings can be leveraged to natively rank at the sub-sentence level, and compare them against specialized models~\cite{chen2023sub} explicitly trained to encode propositions. We note that the encoding here is at sentence-level, unlike in \S{\ref{sec:query_sentence}} where encoding is at passage-level.  Since the marker $m'_q$ in our multi-granular training loss was for sentence-level ranking with passage-level encoding, we use the default marker $m_q$ when ranking at proposition-level with sentence-level encoding. 

\begin{figure*}
    \centering
    \includegraphics[width=1.0\linewidth]{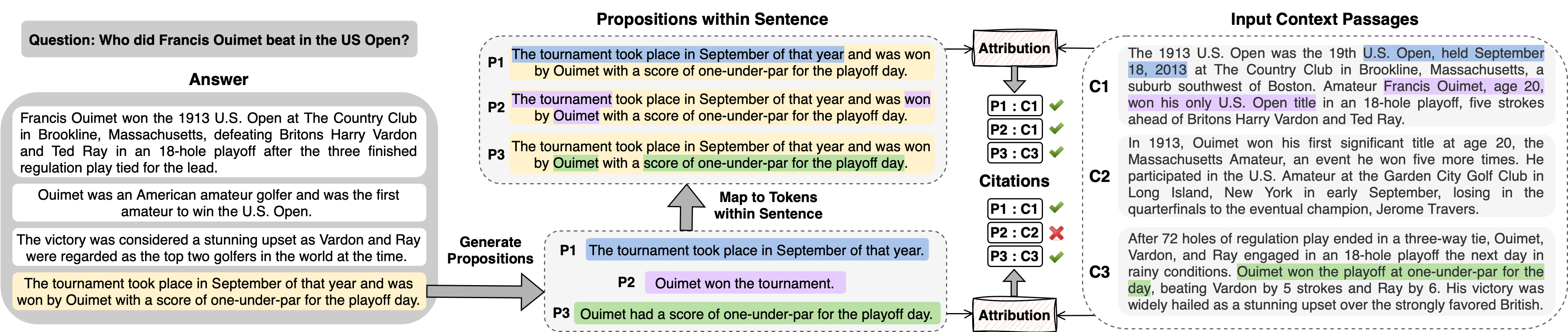}
    \caption{Figure illustrating \textsc{PropCite}, our proposed approach for post-hoc addition of citations to long-form answers. \textsc{PropCite} encodes sentences and uses the propositions \textit{within} them as queries for attribution. 
    The figure shows the propositions highlighted within the current sentence (in yellow), and the corresponding supporting evidence highlighted in the input context passages. \textsc{PropCite} correctly attributes proposition \textit{P2} to context \textit{C1}, while directly encoding and querying using \textit{P2} incorrectly attributes to \textit{C2}.}
    \label{fig:attribution}
\end{figure*}

\subsubsection{Setup}

\paragraph{Datasets}

For the proposition-level ranking evaluation, we use the \textsc{PropSegmEnt}~\cite{chen2023propsegment} dataset, which involves 8.8k propositions as sub-queries for which evidence needs to be obtained from a corpus of 45k human-labeled atomic propositions from 1.5k News or Wikipedia documents in total.

\paragraph{Baselines} For this task, we consider \textsc{SubEncoder}~\cite{chen2023sub} as the primary baseline, a state-of-the-art sub-sentence encoder for proposition-level ranking. \textsc{SubEncoder} has been specifically trained to produce contextual embeddings for atomic propositions in a sentence. Being a single-vector model, \textsc{SubEncoder} produces a single sub-sentence embedding for each atomic proposition in the sentence. We also include other sentence-level embedding approaches, such as GTR~\cite{ni2022large}, Sentence-T5~\cite{ni2022sentence}, as baselines that~\citet{chen2023propsegment} adapt for this task by specifically pooling over the tokens of the proposition to get a single-vector embedding.

\subsubsection{Results}

\begin{table}[t]
    \centering
    \small
    \def\arraystretch{1.3}
    \begin{tabular}{c|cc|cc}
    \multirow{2}{*}{\textbf{Model}} & \multicolumn{2}{c
    |}{\textbf{Proposition}}& \multicolumn{2}{c}{\textbf{Sentence}} \\
    & \textbf{P@1} & \textbf{R@5} & \textbf{P@1} & \textbf{R@5}\\
    \hline
    GTR & 21.9 & 52.5 & 49.4 & 77.0 \\
    ST5 & 26.2 & 57.7 & 50.6 & 79.4 \\
    \textsc{SubEncoder} (GTR) & 40.8 & 72.9 & 42.9 & 82.3 \\
    \textsc{SubEncoder} (ST5) & 41.0 & 72.2 & 43.5 & 81.4 \\
    ColBERTv2 & 46.9 & 74.2 & 54.7 & \textbf{87.8} \\
    \hline
    Ours & \textbf{47.7} & \textbf{74.7} & \textbf{55.0} & 87.4 \\
    \hline
    \end{tabular}
    \caption{Evaluation results on the \textit{Atomic Fact Retrieval} task in \textsc{PropSegmEnt}~\cite{chen2023propsegment}. The encoding level is individual sentences, with each sentence consisting of multiple propositions. All models are based on encoders with 110M parameters. Numbers for GTR, ST5, \textsc{SubEncoder} are from~\citet{chen2023sub}.}
    \label{tab:fact_Attr}
\end{table}

Table \ref{tab:fact_Attr} shows results from the Atomic Fact Retrieval task. First, we observe that the baseline ColBERTv2 already outperforms the state-of-the-art \textsc{SubEncoder} at proposition-level (sub-sentence) ranking. Although our proposed approach adds a sentence-level constrastive loss at passage-level encoding, we do see some improvements when ranking at proposition-level. However, we hypothesize that better proposition-level ranking can be expected by further training with a proposition-level loss in \S{\ref{sec:joint_training}}, which we leave for future work to explore. Nevertheless, given the state-of-the-art performance of multi-vector approaches at proposition-level ranking, we introduce next (in \S{\ref{sec:citation_addition}}) a practical application leveraging this for the task of adding citations to machine-generated text.

\subsection{Sub-Query$\rightarrow$Retrieval Unit Ranking for Citation Addition}
\label{sec:citation_addition}

Retrieval-augmented generation (RAG)~\cite{lewis2020retrieval, izacard2023atlas} produces a long-form answer to a query, given a set of relevant passages as input context. Here, we explore the ability of multi-vector approaches to act as a citation addition approach for attribution in RAG. Specifically, given a set of $K$ passages, and the generated long-form answer, the task involves adding citations to one or more of the input passages, for each sentence in the generated answer. 

We introduce \textsc{PropCite}, a post-hoc citation approach that adds citations to the input context supporting propositions (atomic facts) in the generated text. 
Specifically, \textsc{PropCite} makes use of propositions tagged\footnote{We employ the approach from~\citet{chen2023sub}, which uses a T5 model~\cite{raffel2020exploring} to segment sentences into propositions, that are then converted into token masks by aligning the tokens in each proposition to the sentence.} \textit{within} the generated sentences, so that the corresponding sub-parts can be used as the query to score the input passages and identify the ones that need to be cited. Figure \ref{fig:attribution} illustrates \textsc{PropCite}. Our approach is `post-hoc' since citations are added \textit{after} the text is generated, as opposed to the typical approach of generating text with citations by directly prompting the generation model~\cite{gao2023enabling} to add citations.

\subsubsection{Setup}

\paragraph{Datasets and Metrics} We consider various long-form question answering datasets, specifically ASQA~\cite{stelmakh2022asqa} and ELI5~\cite{fan2019eli5}. The RAG setting involves both $K$=5 and $K$=10 passages as input to the language model to generate the answer. The evaluation of attribution quality is based on the citation precision and recall metrics introduced in~\citet{gao2023enabling}. Citation recall determines if the output is entirely supported by cited passages and citation precision identifies any irrelevant citations. The metrics are computed using TRUE~\cite{honovich-etal-2022-true-evaluating}, a 11B-parameter model trained on a collection of natural language inference datasets, commonly used~\cite{bohnet2022attributed, gao2023rarr} to evaluate attribution by checking whether the cited passages entail the claims in the sentence. 

\paragraph{Baselines} We compare \textsc{PropCite} against the commonly used instruction-driven citation generation~\cite{gao2023enabling}, which we call \textit{Generate}, where the generation model is prompted to output text with citations. We use the same few-shot prompt (provided in appendix) as~\citet{gao2023enabling} to instruct the model to add citations \textit{while} generating the answer. We consider variants of the generation model, a smaller\footnote{We also experimented with Google's Gemma 2B and Microsoft Phi-2 models. Refusal rate was too high for Gemma 2B while Phi-2 had an input context length of only 2048.} 4B Qwen1.5~\cite{bai2023qwen} and a larger 7B Mistral-Instruct~\cite{jiang2023mistral}. We also include comparison with Self-RAG~\cite{asai2023self}, which uses a self-reflective generation framework to adaptively pick input passages to generate from and thereby cite.

\subsubsection{Results}

Table \ref{tab:citation} shows the citation precision (P) and recall (R) numbers comparing the citation quality in the generated text vs our post-hoc \textsc{PropCite} approach. Firstly, the ability to generate text with citations depends heavily on the instruction-following capability of the generation model, with weaker models such as Qwen1.5 4B considerably worse-off compared to Mistral-Instruct 7B. Moreover, even the citation quality of post-hoc approaches depends on the quality of generated text, i.e. when weaker models hallucinate or generate text that cannot be supported by the input context, citation quality is expected to be lower. 

 We observe that \textsc{PropCite} has significantly better citation quality on the 4B and 7B model generations. Even for the Self-RAG models, which were explicitly finetuned for RAG by adding special reflection tokens to cite passsages, we see improvements with \textsc{PropCite}. It is important to note \textsc{PropCite} is post-hoc, and hence can be used with any RAG framework, without needing to tweak the generation model. Moreover, \textsc{PropCite} is light-weight\footnote{While we use a T5 model to explicitly segment sentences into propositions, faster approaches relying on syntactic dependency parsing~\cite{goyal2020evaluating, wanner2024closer} can be a cheaper alternative to get the sub-structures with a sentence that represent the propositions or atomic claims.} and can post-hoc add citations as sentences are generated one-by-one in a streaming setting.

 \begin{table}[t]
    \centering
    \small
    \setlength{\tabcolsep}{0.40em}
    \def\arraystretch{1.3}
    \begin{tabular}{c|c|c|cc|cc}
    \multirow{2}{*}{\parbox{4.5em}{\centering \textbf{Generation Model}}} & \multirow{2}{*}{\textbf{Psg.}} & \multirow{2}{*}{\parbox{3.3em}{\centering \textbf{Citation Method}}} & \multicolumn{2}{c
    |}{\textbf{ASQA}}& \multicolumn{2}{c}{\textbf{ELI5}} \\
    & & &\textbf{P} & \textbf{R} & \textbf{P} & \textbf{R}\\
    \hline
    \multirow{4}{*}{Qwen1.5 4B} & \multirow{2}{*}{5} &\textit{Generate} & 26.9  & 21.3  & 11.0 & 8.6 \\
     &  & \textsc{PropCite}& \textbf{48.9} & \textbf{54.5}  & \textbf{19.5} & \textbf{23.4}\\
     \cline{2-7}
     & \multirow{2}{*}{10} &\textit{Generate} & 14.8  & 11.7  & 5.7 & 4.7\\
     & & \textsc{PropCite}& \textbf{45.3} & \textbf{52.0} & \textbf{18.3} & \textbf{22.9}\\
    \hline
    \multirow{4}{*}{Mistral 7B} & \multirow{2}{*}{5} &\textit{Generate} & 64.9 & 69.5 & 40.5 & 49.0\\
     &  & \textsc{PropCite}& \textbf{65.7} & \textbf{74.2} & \textbf{43.0} & \textbf{51.9}\\
     \cline{2-7}
     & \multirow{2}{*}{10} &\textit{Generate} & 60.2 & 66.7 & 38.0 & 48.8\\
     & & \textsc{PropCite}& \textbf{61.6} & \textbf{71.9} & \textbf{41.9} & \textbf{53.0}\\
     \hline
      \multirow{2}{*}{Self-RAG 7B} & \multirow{4}{*}{5} &\textit{Generate} & 67.9 & 67.1 & - & -\\
     &  & \textsc{PropCite}& \textbf{68.5} & \textbf{68.4} & - & -\\
     \cline{0-0}%
     \cline{3-7}
     \multirow{2}{*}{Self-RAG 13B} &  &\textit{Generate} & 71.4 & 70.5 & - & -\\
     & & \textsc{PropCite}& \textbf{71.6} & \textbf{71.5} & - & -\\
     \hline
    \end{tabular}
    \caption{Table showing precision (P) and recall (R) for different citation addition approaches on the long-form ASQA~\cite{stelmakh2022asqa} and ELI5~\cite{fan2019eli5} question answering datasets. For Self-RAG, we directly use generation outputs from~\citet{asai2023self}.}
    \label{tab:citation}
\end{table}

 \begin{table}[t]
    \centering
    \small
    \setlength{\tabcolsep}{0.45em}
    \def\arraystretch{1.3}
    \begin{tabular}{c|c|c}
     \textbf{Setting}  & \textbf{Precision} & \textbf{Recall} \\
     \hline
     Generate & 64.9 & 69.5 \\
     \hline
     \textsc{PropCite}  & 65.7 & \textbf{74.2} \\
        + \textit{Thresholding} & \textbf{69.2} & 71.1 \\
     \hline
     (i) Propositions as query & 63.5 & 73.9 \\
     (ii) Sentence as query (top 1) & 69.0 & 67.5 \\
     (iii) Sentence as query (top 2) & 51.2 & 72.6\\
     \hline
    \end{tabular}
    \caption{Analysis of citation precision and recall performance on ASQA for Mistral 7B when using top-5 passages as input. We consider different settings, wherein the generated propositions or the sentence itself are used as the query when searching for relevant citations.}
    \label{tab:citation_ablation}
\end{table}

\subsubsection{Analysis}

Table \ref{tab:citation_ablation} shows results for an ablation study with different variants of post-hoc citation addition to demonstrate the benefits of using propositions \textit{within} generated sentences as the query. Firstly, we show numbers for a higher-precision version of \textsc{PropCite} that incorporates thresholding\footnote{To mitigate false positives, we only add a citation if the top-scored passage has a relevance score margin of atleast 1.0.} to decide whether to add a citation for a given proposition in the sentence. Next, we compare against different variants that directly encode the proposition (i) or query using the entire sentence (ii, iii). We can see that directly encoding the proposition, instead of encoding the sentence and using tokens corresponding to the proposition, leads to a drop in precision. This supports our primary hypothesis that encoding at lower-granularity (sentence-level in this case) gives additional context to the token vectors when used for querying at higher granularity (proposition-level in this case). Figure \ref{fig:attribution} illustrates this with an example from the ASQA dataset. \textsc{PropCite} correctly attributes proposition \textit{P2} to input passage \textit{C1} which mentions U.S. Open as the tournament in September that was won by Ouiment. However, directly encoding \textit{P2} misses the context that the tournament occured in September and incorrectly attributes to input passage \textit{C2}, that mentions a different tournament, the Massachusetts Amateur, that Ouiment won.

Moreover, we compare against an alternate approach that just uses the entire sentence as one single query, instead of separately using the individual propositions within the sentences as queries. Tagging the top-1 scored passage as the citation (ii) for that sentence gives a high precision but considerably low recall, while tagging top-2 scored passages as the citations does improve recall but precision suffers a lot. Overall, with \textsc{PropCite}, 66\% of sentences had 1 citation, 30\% had 2 citations and remaining 4\% had more than 2 citations.

\section{Related Work}

The phrase `multi-granularity' can have different meaning depending on the domain in which it is being used. In space of image retrieval, it corresponds to representing different regions of the image separately~\cite{wang2021text, zhang2022multi}. For representation learning, it refers to encoding information at different granularities, i.e. output embedding dimensions, to adapt to the computational constraints of downstream tasks~\cite{kusupati2022matryoshka, li20242d}. Our definition of granularity in text ranking corresponds to ranking relevant sub-units within a given retrieval unit.

Multi-vector approaches~\cite{luan2021sparse, khattab2020colbert, santhanam2022colbertv2} have primarily been used for ranking at the same granularity as the level of encoding, which is typically passage-level. Single vector approaches, on the other hand, inherently do not support ranking at a finer granularity than the level of encoding, thereby needing a separate dense index for each granularity~\cite{chen2023dense}. Hence, specialized models for single-vector embeddings have be introduced for embedding phrases~\cite{lee2021learning}, propositions~\cite{chen2023sub}, sentences~\cite{reimers-2019-sentence-bert} or passages~\cite{karpukhin2020dense}. Our approach leverages multi-vector approaches for ranking at different granularities, while still encoding at a single coarser level of granularity.

Prior approaches that score at different granularities have leveraged custom scoring functions or incorporate separate embeddings.~\citet{chang2023matchacnn} proposes a multi-granularity matching model that uses a convolutional filter for scoring, instead of dot similarity, meaning it cannot be scaled to a retrieval-scale corpus due to the matching function. Hierarchical ranking approaches~\cite{liu2019hierarchical, chu2022h, ma2024multi} also consider multi-granular ranking but require use separate embeddings for each granularity to rank at. In contrast, our approach directly uses multi-vector embeddings from a single-level of encoding to rank at any granularity. Further, our approach use a dot product for scoring at all levels of granularity, meaning the same pre-computed dense corpus index can be used for any granularity.

\section{Conclusion}

 In this work, we introduce \textsc{AGRaME}, which leverages multi-vector embeddings to rank at finer granularities, while encoding is still at a single, coarser level. Our proposed multi-granular contrastive loss for training multi-vector approaches improves sentence ranking performance even with passage-level encoding. We demonstrate that \textsc{AGRaME} can rank at any-granularity, even by using sub-parts of the query for ranking. Leveraging multi-vector approaches' superior performance at proposition-level ranking, our post-hoc attribution approach uses propositions in the generated text to rank input context passages to identify the relevant one to cite. We show superior performance with \textsc{PropCite} over the conventional approach of prompt-driven citation in retrieval-augmented generation.

\section*{Acknowledgement}
 We would like to thank Omar Khattab and members of the Blender NLP group for helpful comments and feedback. We are also grateful to members of the Apple Knowledge Platform team, especially Mostafa Arefiyan, Ihab Ilyas, Theodoros Rekatsinas and Benjamin Han for early discussions. This research is based on work supported by U.S. DARPA KAIROS Program No. FA8750-19-2-1004. The views and conclusions contained herein are those of the authors and should not be interpreted as necessarily representing the official policies, either expressed or implied, of DARPA, or the U.S. Government. The U.S. Government is authorized to reproduce and distribute reprints for governmental purposes notwithstanding any copyright annotation therein.


\bibliography{custom}



\end{document}